\documentclass{article}

 \usepackage[preprint]{neurips_2026}

\usepackage[utf8]{inputenc} 
\usepackage[T1]{fontenc}    
\usepackage{hyperref}       
\usepackage{url}            
\usepackage{booktabs}       
\usepackage{amsfonts}       
\usepackage{nicefrac}       
\usepackage{microtype}      

\usepackage{bbm}
\usepackage{graphicx}
\usepackage{multirow}
\usepackage{makecell}
\usepackage{booktabs}
\newsavebox{\lefttablebox}
\newsavebox{\righttablebox}
\usepackage{amsmath}
\usepackage{pifont}
\usepackage{threeparttable}
\usepackage{colortbl}
\usepackage{float}
\usepackage{pifont}
\usepackage{multirow}
\usepackage{caption}
\definecolor{mygray}{gray}{.92}
\definecolor{ao(english)}{rgb}{0.0, 0.5, 0.0}
\usepackage[dvipsnames,table,xcdraw]{xcolor}

\title{ViewSAM: Learning View-aware Cross-modal Semantics for Weakly Supervised Cross-view Referring Multi-Object Tracking}

\author{%
  Jiawei Ge\\
  Southeast University
  \And
  Xintian Zhang\\
  Southeast University  
  \And 
  Jiuxin Cao\\
  Southeast University
  \And
  Bo Liu\\
  Southeast University
  \And
  Fabian Deuser\\
  Universität der Bundeswehr München
  \And
  Chang Liu\\
  Southeast University
  \And
  Gong Wenkang\\
  Southeast University
  \And
  Siyou Li\\
  Queen Mary, University of London
  \And
  Juexi Shao\\
  Queen Mary, University of London
  \And
  Wenqing Wu\\
  Nanjing University of Science and Technology
  \And
  Chen Feng\\
  Queen's University Belfast
  \And
  Ioannis Patras\\
  Queen Mary, University of London
}

\begin{document}

\maketitle

\begin{abstract}
Cross-view Referring Multi-Object Tracking (CRMOT) aims to track multiple objects specified by natural language across multiple camera views, with globally consistent identities. Despite recent progress, existing methods rely heavily on costly frame-level spatial annotations and cross-view identity supervision. To reduce such reliance, we explore CRMOT under weak supervision by leveraging the capabilities of foundation models. However, our empirical study shows that directly applying foundation models such as SAM2 and SAM3, even with task-specific modifications, fails to accurately understand referring expressions and maintain consistent identities across views. Yet, they remain effective at producing reliable object tracklets that can serve as pseudo supervision. We therefore repurpose foundation models as pseudo-label generators and propose a two-stage framework for weakly supervised CRMOT, using only object category labels as coarse-grained supervision. In the first stage, we design an Affinity-guided Cross-view Re-prompting strategy to refine and associate SAM3-generated tracklets across cameras, producing reliable cross-view pseudo labels for subsequent training. In the second stage, we introduce ViewSAM, a CRMOT model built upon SAM2 that explicitly models view-aware cross-modal semantics. By formulating view-induced variations as learnable conditions, ViewSAM bridges the gap between view-variant visual observations and view-invariant textual expressions, enabling robust cross-view referring tracking with only approximately 10\% additional parameters. Extensive experiments demonstrate that ViewSAM achieves SOTA performance under weak supervision and remains competitive with fully supervised methods.
\end{abstract}

\section{Introduction}

The objective of Cross-view Referring Multi-Object Tracking (CRMOT) \cite{chen2025cross} is to localize and track multiple objects specified by textual descriptions, producing identity-consistent trajectories across time and camera views. By leveraging synchronized observations from multiple cameras, CRMOT extends single-view Referring Multi-Object Tracking (RMOT) \cite{wu2023referring} to multi-camera scenarios and exploits cross-view complementarities to alleviate long-standing challenges in RMOT, such as severe occlusions and target disappearance. This capability is essential for enabling language-driven perception systems in the real world, particularly autonomous driving ~\cite{fischer2022cc} and embodied AI~\cite{ziliotto2025tango}.

Despite its importance, existing CRMOT methods \cite{hao2024divotrack,chen2025cross,gao2023multi,zhang2025dual,zhen2024gmt,fan2025all} are predominantly developed under a fully supervised learning paradigm. These approaches rely on large-scale datasets annotated with dense frame-level spatial labels and cross-view identity correspondences. However, collecting such fine-grained annotations is extremely expensive, as the labeling cost grows rapidly with both the number of frames and views, making it difficult to scale CRMOT solutions to open-world scenes.

To alleviate such reliance, we explore a new learning paradigm termed \textbf{Weakly Supervised Cross-view Referring Multi-Object Tracking (WSCRMOT)}, inspired by recent advances in weakly supervised learning \cite{ge2024consistencies,xu2014large,zhu2024weaksam}. In this setting, the training data provide only \textbf{coarse-grained object category labels and raw multi-view videos}~\cite{zhou2018brief}, without any spatial annotations, cross-view identities, or referring expressions. While such weak supervision greatly reduces annotation costs, it also introduces a fundamental challenge: \emph{how can a model learn cross-view referring tracking without explicit spatial or identity supervision?}

Recent progress in large-scale foundation models~\cite{brown2020language,radford2021learning,kirillov2023segment} provides a promising direction. Models such as SAM~\cite{kirillov2023segment} and its successors SAM2~\cite{ravisam} and SAM3~\cite{carion2025sam} exhibit remarkable generalization capabilities in segmentation and tracking, and have increasingly been explored as external supervision sources for weakly supervised learning~\cite{zhu2024weaksam,kweon2024sam,he2023weakly}. With their flexible prompting mechanisms and strong capabilities, SAM2 and SAM3 provide a natural starting point for exploring WSCRMOT.

\begin{figure}[t]
\centering 
\includegraphics[width = 0.94\linewidth]{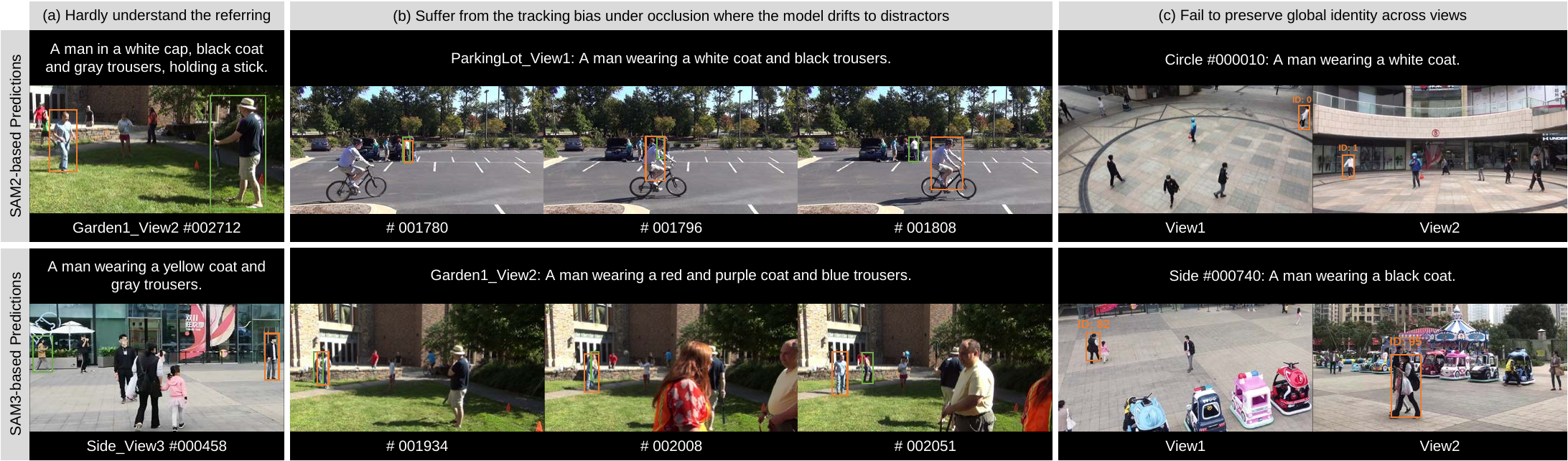}
\caption{Empirical visualizations (\textcolor{orange}{Prediction} and \textcolor{ao(english)}{GT}): (a) hard to understand referring, (b) drift to distractors under occlusion, and (c) fail to preserve cross-view ID with object feature clustering.}
\label{empirical}
\end{figure}

However, our empirical study (Fig.~\ref{empirical}) reveals key limitations when applying SAM-based models in WSCRMOT. First, they lack robust semantic understanding for language-guided long-term tracking, often suffering from the \textbf{tracking bias} where the model drifts to distractors that partially match the textual prompt. Moreover, they fail to maintain global identity consistency, as the same object may exhibit substantial variations in appearance and motion patterns across views. Taken together, these observations indicate that, while SAM-based models are not well suited for direct application in this task, they can instead be repurposed as effective pseudo-label generators under weak supervision.

\begin{figure}[b]
\centering 
\includegraphics[width = 0.96\linewidth]{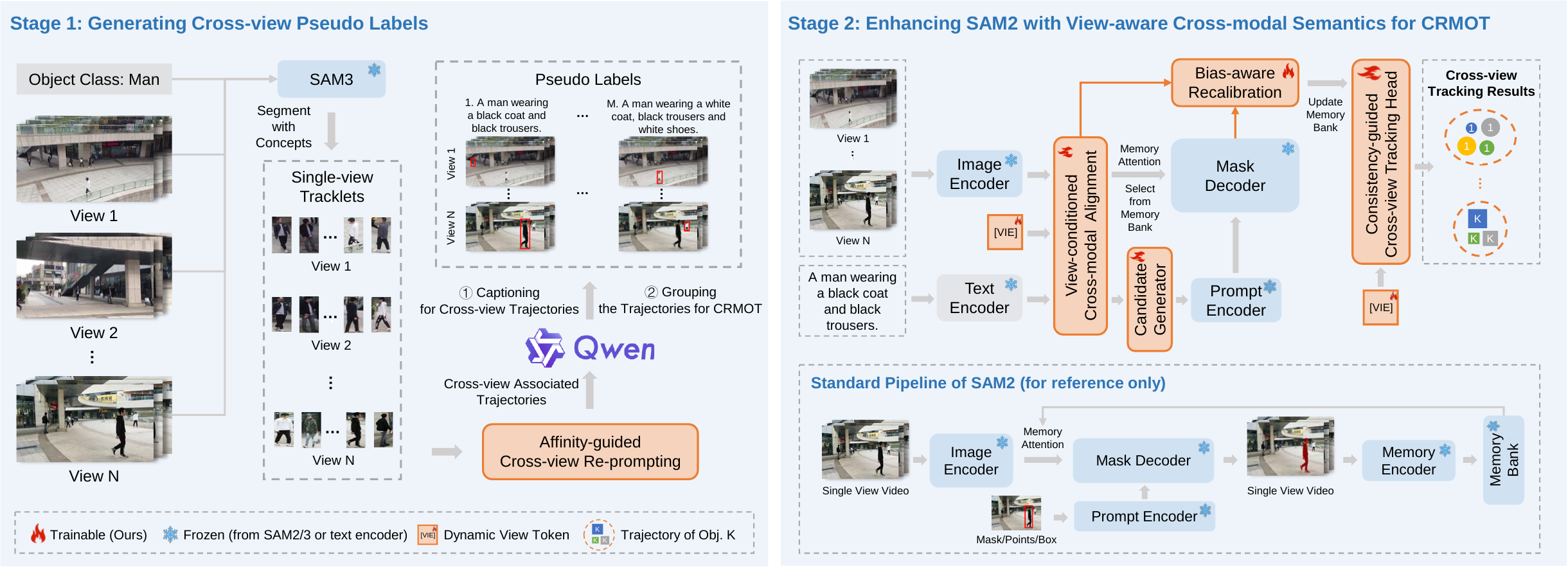}
\caption{Overview of our two-stage framework for WSCRMOT. In Stage 1, we generate pseudo labels by refining and associating SAM3 tracklets with our Affinity-guided Cross-view Re-prompting strategy. In Stage 2, limited extra parameters are introduced to enhance SAM2 with view-aware cross-modal semantics modeling for CRMOT, where view-induced variations are treated as learnable conditions rather than detrimental factors. The standard SAM2 pipeline is also shown for reference.
}
\label{overview}
\end{figure}

Building on these insights, we propose a two-stage framework for Weakly Supervised Cross-view Referring Multi-Object Tracking, as illustrated in Fig. \ref{overview}. In the first stage, we exploit the strong tracking capability of foundation models to generate cross-view pseudo labels. Starting from single-view tracklets produced by SAM3, we design an \textbf{Affinity-guided Cross-view Re-prompting} strategy that iteratively refines object predictions and aligns tracklets across views. By integrating these refined tracklets with trajectory-level descriptions generated by a multimodal large language model (MLLM), we obtain high-quality pseudo supervision for downstream training.

In the second stage, we introduce \textbf{ViewSAM}, a SAM2-based CRMOT model that explicitly learns view-aware cross-modal semantics throughout single-view tracking and cross-view association. Instead of treating view variations as factors to be ignored or eliminated, ViewSAM formulates them as learnable semantic conditions. Specifically, we propose a View-conditioned Cross-modal Alignment module to comprehend referring expressions, jointly reasoning over visual and textual modalities through interactions with a learnable dynamic View Token. To mitigate the tracking bias mentioned earlier, we further propose a Bias-aware Recalibration module that encourages the model to refocus on objects that are more consistent with the referring expression. For cross-view association, we further devise a Consistency-guided Cross-view Tracking Head that leverages the learned dynamic view token to modulate tracklet representations and suppress view-induced discrepancies. Together with consistency-guided objectives, this mechanism projects tracklets into a view-invariant space and provides explicit supervision for cross-view tracklet association. Through these designs, ViewSAM bridges the gap between view-variant visual appearances and view-invariant textual expressions, enabling robust cross-view referring tracking. To sum up, our main contributions are as follows:

\begin{enumerate}

\item To the best of our knowledge, we propose the first weakly supervised framework for Cross-view Referring Multi-Object Tracking that substantially reduces the reliance on dense annotations, establishing a solid baseline for future research.

\item We propose an Affinity-guided Cross-view Re-prompting strategy to generate reliable cross-view pseudo labels, which models cross-view affinity for consistent identity association and refines trajectories iteratively.

\item We propose ViewSAM, a CRMOT model built upon SAM2 that explicitly models view-aware cross-modal semantics, achieving robust referring understanding and consistent cross-view tracking with only \textasciitilde 10\% additional parameters.

\item Extensive experiments demonstrate that the proposed framework achieves SOTA performance under weak supervision, substantially narrowing the gap to fully supervised methods.

\end{enumerate}

\section{Related Work}

\subsection{Referring Object Tracking}

Referring Object Tracking (ROT) aims to localize and track objects in videos according to explicit referring signals such as natural language descriptions or visual prompts. Early studies mainly focus on the single-object setting, often referred to as Vision–Language Tracking (VLT) \cite{li2017tracking,wang2021towards}, where visual features are aligned with textual embeddings via cross-modal interaction \cite{zhou2023joint,guodivert,ge2024consistencies,wang2025r1trackdirectapplicationmllms,ge2025beyond}. 

Recent work extends this paradigm to multi-object scenarios, known as Referring Multi-Object Tracking (RMOT) \cite{wu2023referring}. Existing RMOT methods can be broadly categorized into two-stage approaches \cite{du2024ikun,li2025lamot,li2025language} and end-to-end architectures \cite{wu2023referring,xiao2025temporal,zhuang2025cgatracker,liang2025cognitive,li2025visual}. However, these methods are limited to single-view settings. To handle multi-camera scenarios, several studies investigate cross-view multi-object tracking by associating identities across cameras using appearance and motion cues \cite{hao2024divotrack,gao2023multi,zhang2025dual,zhen2024gmt,fan2025all}. Building upon it, Chen et al.~\cite{chen2025cross} introduce Cross-view Referring Multi-Object Tracking (CRMOT), where objects specified by language expressions are consistently tracked across synchronized camera views. Unlike them, we explore CRMOT under a weakly supervised setting.

\subsection{Segment Anything Models}

The Segment Anything Model (SAM) \cite{kirillov2023segment} is a powerful vision foundation model for generic object segmentation with strong zero-shot generalization. Its promptable design has made it widely adopted as a universal segmentation backbone or pseudo-label generator for downstream tasks. To extend SAM to videos, SAM2 \cite{ravisam} introduces memory mechanisms for mask propagation across frames. Subsequent works further enhance its tracking robustness \cite{yang2024samurai,ding2025sam2long,jiang2025sam2mot}. More recently, SAM3 \cite{carion2025sam} integrates a DETR-based concept detector to segment all instances of a given concept prompt. 

However, SAM-based models are not designed to understand complex referring expressions or maintain identity consistency across views. In this work, we leverage the strong generalization ability of SAM models to generate pseudo supervision and propose ViewSAM, which explicitly models view-induced variations as learnable semantic conditions for robust cross-view referring tracking.

\section{The Proposed Method} 

\noindent
\textbf{Problem Setting.}
Given a set of synchronized multi-view video sequences 
$\mathcal{V}= \left\{\mathcal{V}_1, \mathcal{V}_2, \ldots, \mathcal{V}_{N}\right\}$ 
captured from $N$ cameras observing the same scene, and a referring expression $\mathcal{R}$, the goal of cross-view referring multi-object tracking (CRMOT) is to detect and track all objects specified by $\mathcal{R}$. The output is a set of object trajectories, where each target object is assigned a global ID across views.

Unlike fully supervised CRMOT methods that require referring expressions, bounding box annotations, and cross-view identity labels, we consider a weakly supervised setting in this work where only the object category label $\mathcal{C}$ and raw synchronized multi-view videos $\mathcal{V}$ are available during training. 

\begin{figure*}[t]
    \centering
    \includegraphics[width=\linewidth]{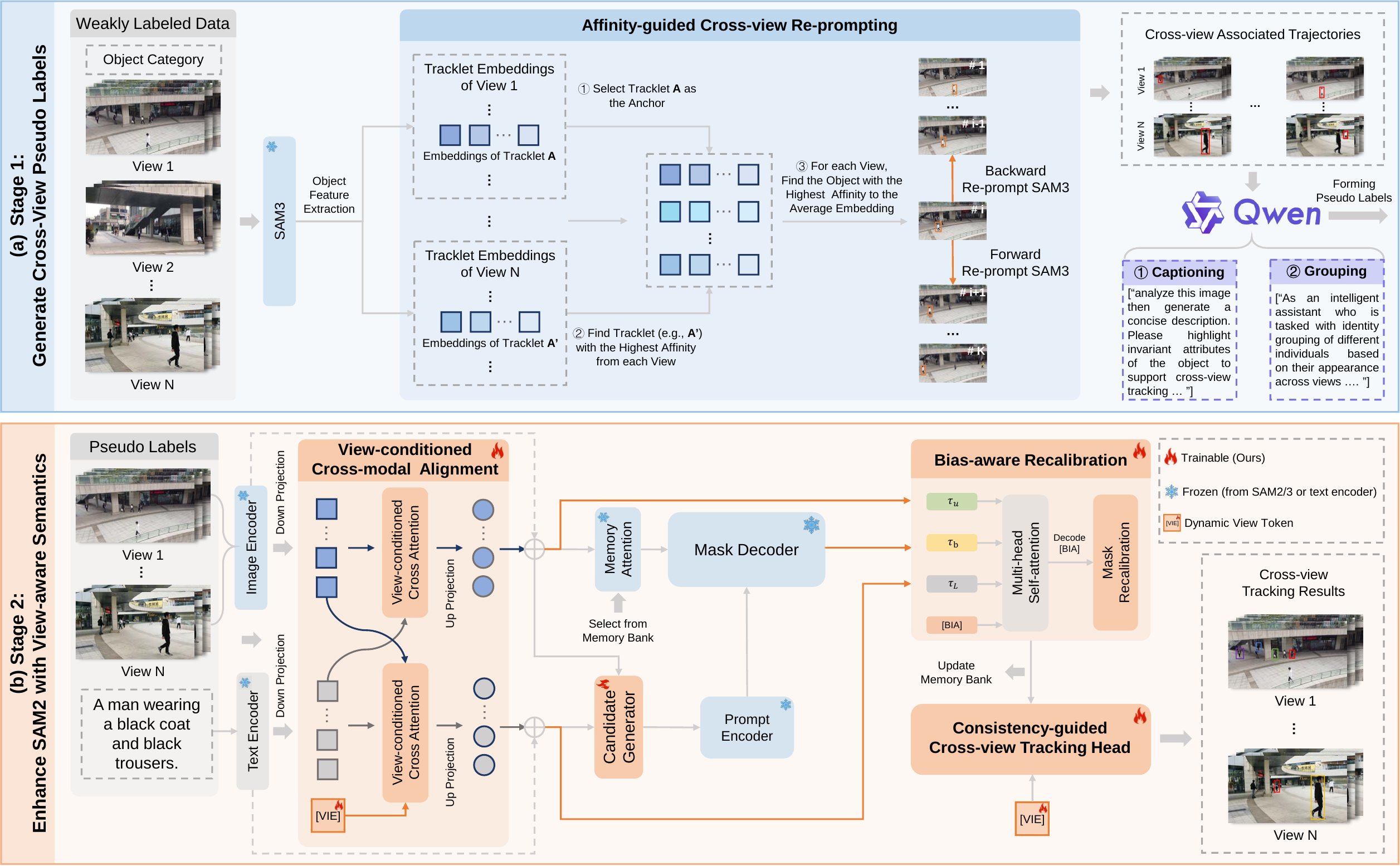}
    \caption{The overall pipeline of our framework. (a) Affinity-guided Cross-view Re-prompting generates cross-view pseudo labels from weakly labeled multi-view videos by extracting tracklets with SAM3, aligning high-affinity trajectories across views, and refining them through forward and backward re-prompting. (b) Using these pseudo labels, we enhance SAM2 with view-aware cross-modal semantics, including a View-conditioned Cross-modal Alignment module that reasons over visual and textual features, followed by the Bias-aware Recalibration to overcome the tracking bias and a Consistency-guided Cross-view Tracking Head to produce the trajectories with global IDs.}
    \label{framework}
\end{figure*}

\noindent
\textbf{Overview.}
As shown in Fig.~\ref{framework}, our framework consists of two stages: In the first stage (Sec.~\ref{s3-2}), we leverage SAM3's concept-level tracking capability to generate cross-view pseudo labels via an Affinity-guided Cross-view Re-prompting strategy. In the second stage (Sec.~\ref{s3-3}), we introduce ViewSAM that incorporates view-aware cross-modal semantics and enables ID-consistency across views. Before elaborating them, we briefly review the architectures of SAM2 and SAM3 in Sec.~\ref{s3-1}.

\subsection{Background on SAM2 and SAM3}
\label{s3-1}

SAM2~\cite{ravisam} is a foundation model for video object segmentation composed of an Image Encoder, a Prompt Encoder, a Mask Decoder, and a Memory Mechanism for temporal mask propagation.

\noindent
\textbf{Memory Mechanism.}
For each frame, the Image Encoder extracts visual tokens, which interact with historical memory tokens from Memory Bank. Through memory attention, the current frame features attend to previous frames to incorporate temporal context. After mask prediction, a Memory Encoder converts current features and predicted masks into new memory tokens.

\noindent
\textbf{Prompt Encoding and Mask Decoding.}
SAM2 supports sparse prompts (points or boxes) and dense prompts (masks). 
Sparse prompts are encoded using positional and type embeddings, while dense prompts are fused with image features. 
The Mask Decoder employs a two-way transformer to decode prompt-conditioned features for initial frame and memory-conditioned features for the subsequent.

SAM3~\cite{carion2025sam} extends SAM2 toward concept-level segmentation by introducing a \textbf{heavy} concept-aware detector while \textbf{keeping the underlying video tracking backbone unchanged.} Yet, it is not designed to interpret complex referring expressions. Hence, we adopt SAM2 as the base tracker for efficiency.

\subsection{Cross-view Pseudo Label Generation}
\label{s3-2}

Since the training data provide only category labels, directly learning a CRMOT model is challenging due to the absence of object locations and cross-view identity annotations. To address this, we first leverage SAM3 to localize candidate objects in each view using the category label as a text prompt. We then associate tracklets across views using an Affinity-guided Cross-view Re-prompting strategy. Based on these preliminary cross-view single-object tracklets, we further refine and expand them into pseudo labels for CRMOT through the LLM-driven Referring-level Multi-object Grouping.

\subsubsection{Affinity-guided Cross-view Re-prompting.}

Our strategy mainly consists of two components: the Affinity-guided Cross-view Association and the Bi-directional Re-prompting. By leveraging the embedding affinity as a semantic constraint, we iteratively establish and refine cross-view correspondences for each object.

Formally, for each camera view $i$, we first prompt SAM3 with the category label $\mathcal{C}$ to segment candidate object instances, producing a set of single-view candidate tracklets:
$
\mathcal{T}_i = \{\mathcal{T}_i^1, \mathcal{T}_i^2, \dots, \mathcal{T}_i^{|\mathcal{T}_i|}\}, 
\quad i = 1,\dots,N. 
$
Each tracklet $\mathcal{T}_i^k$ is a temporally ordered mask sequence corresponding to object $k$:
$
\mathcal{T}_i^k = \{M_{t,i}^k\}_{t\in\Gamma_i^k},
$
where $\Gamma_i^k$ denotes the set of valid frames containing object $k$ within $\mathcal{T}_i^k$.

Then, we compute a tracklet-level embedding by averaging frame-wise object features:
\begin{equation}
E_i^k=
\frac{1}{|\Gamma_i^k|}
\sum_{t\in\Gamma_i^k}f_\theta(I_{t,i}, M_{t,i}^k),
\qquad
E_i^k \in \mathbb{R}^{d_{\mathrm{id}}},
\end{equation}
where $I_{t,i}$ is image at frame $t$ and view $i$, and $f_\theta(\cdot,\cdot)$ denotes an off-the-shelf ReID head \cite{zhou2019omni} that extracts a $d_{\mathrm{id}}$-dimensional embedding.

\noindent
\textbf{Affinity-guided Cross-view Association.}
Given an anchor tracklet $\mathcal{T}_i^k$, we compute cosine similarities between its embedding $E_i^k$ and all candidate tracklet embeddings from every other view $j\neq i$. For each target view, we retain the top-affinity candidate only if its similarity exceeds a reliability threshold; otherwise that view is treated as unmatched. In this way, we form a preliminary cross-view tracklet set for object $k$. Since this stage serves only as an initialization, duplicated matches across different anchors are allowed and are later corrected by the bi-directional re-prompting strategy.
\noindent
\textbf{Bi-directional Re-prompting for Identity Refinement.}
The preliminary association can still be noisy due to occlusion, appearance variations, or distractors. We further refine it by establishing an aggregated identity prototype from the confidently matched tracklets, which aims to capture semantic characteristics shared across views while suppressing frame-level noise:
\begin{equation}
\bar E^k = \frac{1}{|\mathcal{A}^k|}\sum_{(i',k')\in\mathcal{A}^k} E_{i'}^{k'},
\end{equation}
where $\mathcal{A}^k$ denotes the set of matched tracklets retained for the current anchor after affinity filtering. For each view, we search over the candidate object masks predicted by SAM3, and select the mask whose embedding has the highest affinity with the prototype $\bar E^k$. This selected mask is used as a renewed mask prompt to initialize SAM3 tracking in both forward and backward temporal directions within that view, producing a refined trajectory with improved temporal consistency. Finally, we obtain a refined set of cross-view single-object trajectories, serving as candidates for pseudo labels.

\subsubsection{LLM-driven Referring-level Multi-object Grouping.}

While the previous step establishes cross-view correspondences for individual objects, CRMOT still requires supervision at the referring level, where one expression corresponds to multiple objects. To this end, we leverage the MLLM (Qwen3-VL-8B \cite{bai2025qwen3}) to generate attribute-centric descriptions for each trajectory, capturing appearance cues such as color, clothing, and accessories. These descriptions are embedded into a shared semantic space using the MLLM encoder. Based on the semantic similarity, trajectories are then grouped under the same referring expression. Finally, each referring expression with grouped cross-view trajectories, forms CRMOT pseudo labels.

\subsection{Enhance SAM2 with View-aware Cross-modal Semantics}
\label{s3-3}
As illustrated in Fig.~\ref{framework}(b), given multi-view video streams and a referring expression, ViewSAM first aligns visual features with textual embeddings using a \textbf{View-conditioned Cross-modal Alignment (VC-CMA)} module guided by the learnable dynamic View Token. Then the view-aware cross-modal representations are fed into the Candidate Generator to produce box prompts that guide mask decoding in SAM2, extending it to multi-object localization. Afterwards, a \textbf{Bias-aware Recalibration (BAR)} module dynamically shifts the model's focus on objects that better match the referring expression.

For cross-view association, the predicted tracklets are processed by a \textbf{Consistency-guided Cross-view Tracking Head}, which leverages the learned dynamic view token to modulate tracklet representations before association. Under consistency-guided objectives, the tracklet features are projected into a shared view-invariant embedding space, enabling reliable cross-view identity alignment. All the details about the architecture and training losses for each component are provided in \ref{main-components}.

\subsubsection{View-conditioned Cross-modal Alignment}

To mitigate view-induced variations in referring expression comprehension, we introduce a learnable dynamic View Token built upon the popular Adapter framework \cite{houlsby2019parameter}. Through interactions with visual and textual features, the dynamic View Token enhances multi-modal representations, enabling VC-CMA to learn view-aware cross-modal representations for robust target localization across views. \noindent
\textbf{Learnable Dynamic View Token.}
For each view $i$ with a clip of $T$ frames, we construct a dynamic view-conditioned token, where a learnable base embedding $g_i \in \mathbb{R}^d$ captures static view priors:
\begin{equation}
\hat e_{t,i} =
\mathrm{Norm}\!\left(
g_i + \lambda_{\mathrm{dyn}}\,
\mathrm{Norm}\!\left(
W_{\mathrm{view}}\,
\mathrm{Pool}_{\mathrm{avg}}(F_{t,i})
\right)
\right),
\end{equation}
where $F_{t,i} \in \mathbb{R}^{d \times H \times W}$ denotes the visual feature map at frame $t$ and view $i$, $W_{\mathrm{view}} \in \mathbb{R}^{d\times d}$ is a linear projection, and $\lambda_{\mathrm{dyn}}$ controls the contribution of the frame-level visual context. For temporal stability, we smooth the view token using an exponential moving average (the decay factor $\alpha \in [0,1]$):
\begin{equation}
e_{t,i} =
\mathrm{Norm}\!\left(
\alpha e_{t-1,i} + (1-\alpha)\hat e_{t,i}
\right),
\qquad
e_{1,i}=\hat e_{1,i}.
\end{equation}
\noindent
\textbf{View-conditioned Cross Attention.}
Let $T_{\mathrm{text}} \in \mathbb{R}^{M\times d}$ denote textual tokens and $X_{t} \in \mathbb{R}^{HW\times d}$ denote visual tokens obtained from the feature map $F_{t,i}$. Following the Adapter's design \cite{houlsby2019parameter}, tokens are projected into a bottleneck space via $X_t^s = X_t W_v^{\downarrow}$ and $T_{\mathrm{text}}^s = T_{\mathrm{text}} W_t^{\downarrow},$ with dimension $d_s < d$.

To bridge the gap between view-variant visual observations and view-invariant textual expressions, we inject view-related prior into query and key branches via view-conditioned cross attention (VCCA):
\begin{equation} \mathrm{VCCA}(U,R;e) = U \odot \mathrm{MHA}(U + b_u(e),\, R + b_r(e),\, R),\quad b_u(e)=eW_u, b_r(e)=eW_r, \end{equation}
where $U$ denotes the query tokens and $R$ denotes the reference tokens (serving as key and value in standard cross-attention). Here, the Hadamard product $\odot$ acts as a lightweight feature-wise gating mechanism, enabling view-aware conditioning while preserving the original token semantics.

We then update visual tokens using the frame-level view token and project them to the original space:
\begin{equation}
X_{\mathrm{ca}}^t =
X_t +
\mathrm{VCCA}(X_t^s, T_{\mathrm{text}}^s; e_{t,i})
W_v^{\uparrow}.
\end{equation}
For textual tokens, we instead use a clip-level visual summary
$\bar X^s$ and a clip-level view token $e_i^{\mathrm{clip}}$:
\begin{equation}
T_{\mathrm{ca}} =
T_{\mathrm{text}} +
\mathrm{VCCA}(T_{\mathrm{text}}^s, \bar X^s; e_i^{\mathrm{clip}})
W_t^{\uparrow}, \quad \bar X^s = \frac{1}{T}\sum_{t=1}^{T} X_t^s,e_i^{\mathrm{clip}} = \frac{1}{T}\sum_{t=1}^{T} e_{t,i}.
\end{equation}
Finally, these view-aware representations are fed into the Candidate Generator to produce box prompts that highlight regions likely to contain targets, which, together with the representations, guide mask decoding in SAM2, hence extending it from single-object to multi-object localization.

\subsubsection{Bias-aware Recalibration}

SAM-based trackers suffer from the aforementioned tracking bias, where memory-guided decoding amplifies historical errors and causes drift under occlusion. To mitigate this, we propose Bias-aware Recalibration (BAR), which leverages memory-free features to detect failures and re-align predictions. Let $F_{t,i}^{u}$ and $F_{t,i}^{b}$ denote pre- and post-memory-attention features (i.e., un-biased vs. biased), with converted soft masks $M_{t,i}^{k,s}$ for $s \in \{u,b\}$. Their object-level tokens are then extracted to form the referring-aware representations:
\begin{equation}
o_s^k = \mathrm{Norm}\!\left(
\frac{\sum_x F_{t,i}^s(x)\,M_{t,i}^{k,s}(x)}
{\sum_x M_{t,i}^{k,s}(x)}
\right),\;
\alpha_s^k = \mathrm{softmax}_k\!\left(\langle o_s^k, t_{\mathcal{R}} \rangle\right),\;
\tau_s = \mathrm{Norm}\!\left(\sum_k \alpha_s^k o_s^k\right)
\end{equation}
Here $t_{\mathcal{R}}$ is the referring token derived from textual tokens $T_{\mathrm{text}}$. Afterwards, we introduce a learnable token $[\mathrm{BIA}]$ to assess whether the current predictions are reliable through the self-attention operation $\mathrm{SA}(\cdot)$. Finally, we adaptively fuse the two predictions based on the resulting bias score $p_{\mathrm{bias}} \in [0,1]$:
\begin{equation}
p_{\mathrm{bias}} = \phi\!\left(\mathrm{SA}([\mathrm{BIA}], t_{\mathcal{R}}, \tau_u, \tau_b)\right), \quad M_{t,i}^{k} = (1 - p_{\mathrm{bias}})\, M_{t,i}^{k,b} + p_{\mathrm{bias}}\, M_{t,i}^{k,u}.
\end{equation}
\subsubsection{Consistency-guided Cross-view Tracking Head}
We propose Consistency-guided Cross-view Tracking (CGCT) head, which is mainly based on CNNs, to learn a view-invariant embedding space for cross-view identity association. It explicitly enforces identity consistency across time and views via view-conditioned modulation and consistency-guided objectives. Given frame $t$ in view $i$, let $\{F_{t,i}^{\ell}\}_{\ell=1}^{L}$ denote multi-scale features. We first perform masked average pooling over multi-scale features for each candidate mask $M_{t,i}^{k}$ and fuse them:
\begin{equation}
f_{t,i}^{k}
=
\operatorname{Norm}\!\left(
\operatorname{Fuse}\Big(
\{\operatorname{Pool}(F_{t,i}^{\ell}, M_{t,i}^{k})\}_{\ell=1}^{L}
\Big)
\right),
\end{equation}
where $\operatorname{Pool}(\cdot)$ denotes masked average pooling, and $\operatorname{Fuse}(\cdot)$ is concatenation followed by an MLP.
\noindent
\textbf{View-conditioned Modulation.}
We incorporate the dynamic view token $e_{t,i}$ via Feature-wise Linear Modulation \cite{perez2018film} controlled by modulation strength $\delta$, mitigating view-specific appearance variations:
\begin{equation}[\gamma_{t,i}, \beta_{t,i}]
=
\operatorname{MLP}(e_{t,i}), \;
z_{t,i}^{k}
=
\operatorname{Norm}\left((1 + \delta \tanh(\gamma_{t,i}))\odot f_{t,i}^{k}
+
\delta \beta_{t,i}
\right), \;
\gamma_{t,i}, \beta_{t,i} \in \mathbb{R}^d
\end{equation}
\noindent
\textbf{Consistency-guided Objectives.}
Let $\mathcal{T}_g^i$ denote the trajectory of identity $g$ within view $i$, $\bar z_g^i$ the view-specific prototype, and $\bar z_g$ the global identity prototype aggregated across views. The prototypes are computed as the mean embeddings over instances. Here, we formulate a two-level objective:
\begin{equation}
\small
\begin{aligned}
\mathcal{L}_{\mathrm{CGCT}}
&=
\sum_{g}
\Bigg[
\underbrace{
\sum_{i}\sum_{(t,k)\in\mathcal{T}_g^i}
\left\|
z_{t,i}^{k} - \bar z_g^i
\right\|_2^2
}_{\text{intra-consistency}}
+
\lambda
\underbrace{
\sum_{i}
\left\|
\bar z_g^i - \bar z_g
\right\|_2^2
}_{\text{inter-consistency}}
\Bigg],
\end{aligned}
\end{equation}
where the intra-consistency term enforces temporal coherence within each view by compacting trajectory embeddings around the view-specific prototype, while the inter-consistency term aligns view-specific prototypes to a shared global identity prototype for cross-view association.

\section{Experiments}

\subsection{Experiment Setup}

\noindent
\textbf{Dataset and Metrics.}
For evaluation, we conduct experiments on the CRTrack~\cite{chen2025cross} benchmark for CRMOT and follow its evaluation protocol with CVR-IDF1 (cross-view referring IDF1) and CVR-MA (cross-view referring matching accuracy), which measure identity consistency across time and views, and correctness of cross-view identity matching, under referring constraints, respectively.

\noindent
\textbf{Implementation Details.}
We use RoBERTa \cite{liu2019roberta} (124.7M) as the text encoder and SAM2 (80.8M) as the base video tracker, both of which are frozen. Only the VC-CMA (\textasciitilde 4.6M), Candidate Generator (\textasciitilde 12.1M), BAR (\textasciitilde 1.1M), and CGCT (\textasciitilde 2.3M) are trained (\textasciitilde 10\% extra parameters). Under the weakly supervised setting, the Candidate Generator is distilled from APTM \cite{yang2023towards} and trained on ID-agnostic single-view pseudo labels (lr: 4e-5, 12 epochs). VC-CMA and BAR are pre-trained on Ref-YoutubeVOS \cite{seo2020urvos} (lr: 2e-4, 8 epochs), and then fine-tuned on ID-agnostic cross-view pseudo labels (lr: 2e-5, 20 epochs). CGCT is trained on refer-agnostic cross-view pseudo labels (lr: 2e-4, 10 epochs). For inference, clip length $T$ is 8. All experiments are conducted on 2 NVIDIA A800 GPUs.

\begin{table*}[t]
\centering

\caption{
Performance comparison of SOTA methods on the CRTrack benchmark. SAM2 and SAM3 are adapted with a cosine-based clustering for cross-view association. Best results within each setting are highlighted in \textbf{bold}. For brevity, metric names are \textbf{simplified} here by removing ``CVR-'' prefix.
}
\setlength{\tabcolsep}{3.2pt}
\renewcommand{\arraystretch}{0.9}
\begin{tabular}{@{} l l l |ll| ll ll ll @{}}
\toprule
\multicolumn{11}{c}{\textbf{In-domain Evaluation}} \\
\midrule

\multirow{2}{*}{Method} 
& \multirow{2}{*}{Setting} 
& \multirow{2}{*}{Publication}
& \multicolumn{2}{c|}{All}
& \multicolumn{2}{c}{Circle}
& \multicolumn{2}{c}{Gate2}
& \multicolumn{2}{c}{Side} \\

\cmidrule(lr){4-5}\cmidrule(lr){6-7}\cmidrule(lr){8-9}\cmidrule(lr){10-11}

& & 
& IDF1 & MA
& IDF1 & MA
& IDF1 & MA
& IDF1 & MA \\

\midrule

TransRMOT \cite{wu2023referring} & \multirow{3}{*}{Fully} & CVPR23
& 23.30 & 8.03
& 18.85 & 6.94
& 68.03 & 28.51
& 14.33 & 2.65 \\


CRTracker~\cite{chen2025cross} &  & AAAI25
& 54.88 & 35.97
& 58.38 & 42.44
& \textbf{91.60} & 73.40
& 37.97 & 14.87 \\

\textbf{ViewSAM} &  & \textbf{Ours}
& \textbf{57.73} & \textbf{43.72}
& \textbf{60.84} & \textbf{48.85}
& {89.21} & \textbf{81.39}
& \textbf{43.02} & \textbf{24.33} \\

\midrule

SAM2~\cite{ravisam} & \multirow{2}{*}{Zero-shot} & ICLR25
& 3.05 & 0.28
& 1.69 & 0.16
& 8.53 & 0.52
& 3.05 & 0.35 \\

SAM3~\cite{carion2025sam} &  & ICLR26
& \textbf{13.44} & \textbf{2.89}
& \textbf{10.52} & \textbf{2.62}
& \textbf{32.40} & \textbf{8.78}
& \textbf{11.02} & \textbf{1.30} \\

\midrule

CRTracker~\cite{chen2025cross}  & \multirow{2}{*}{Weakly} & AAAI25
&33.06  & 21.40
& 31.54 & 20.25
& 71.86 & 57.10
& 22.15  & 11.03  \\

\textbf{ViewSAM} &  & \textbf{Ours}
& \textbf{38.80} & \textbf{26.95}
& \textbf{35.59} & \textbf{27.66}
& \textbf{72.48} & \textbf{61.56}
& \textbf{31.86} & \textbf{14.47} \\

\midrule
\multicolumn{11}{c}{\textbf{Cross-domain Evaluation}} \\
\midrule

\multirow{2}{*}{Method} 
& \multirow{2}{*}{Setting} 
& \multirow{2}{*}{Publication}
& \multicolumn{2}{c|}{All}
& \multicolumn{2}{c}{Garden1}
& \multicolumn{2}{c}{Garden2}
& \multicolumn{2}{c}{ParkingLot} \\

\cmidrule(lr){4-5}\cmidrule(lr){6-7}\cmidrule(lr){8-9}\cmidrule(lr){10-11}

& & 
& IDF1 & MA
& IDF1 & MA
& IDF1 & MA
& IDF1 & MA \\

\midrule

TransRMOT \cite{wu2023referring} & \multirow{3}{*}{Fully} & CVPR23
& 3.66 & 0.20
& 2.85 & 0.01
& 4.23 & 0.55
& 3.87 & 0.00 \\


CRTracker~\cite{chen2025cross} &  & AAAI25
& 12.52 & 2.32
& 14.96 & 2.77
& 11.87 & 2.80
& 10.66 & 1.30 \\

\textbf{ViewSAM} &  & \textbf{Ours}
& \textbf{15.86} & \textbf{8.35}
& \textbf{19.51} & \textbf{10.08}
& \textbf{14.22} & \textbf{7.98}
& \textbf{13.83} & \textbf{6.95} \\

\midrule

SAM2~\cite{ravisam} & \multirow{2}{*}{Zero-shot} & ICLR25
& 7.78   & 1.81
& \textbf{13.93} & \textbf{5.06}
& 4.23 & 0.25
& 5.17 & 0.13 \\

SAM3~\cite{carion2025sam} &  & ICLR26
& \textbf{8.00} & \textbf{1.93}
& 8.58 & 3.44
& \textbf{6.90} & \textbf{1.06}
& \textbf{8.63} & \textbf{1.29} \\

\midrule
CRTracker~\cite{chen2025cross}  & \multirow{2}{*}{Weakly} & AAAI25
&8.12  &1.95 
&9.02  &1.76 
&7.68  &2.19 
&7.65  &1.88  \\

\textbf{ViewSAM} &  & \textbf{Ours}
& \textbf{11.48} & \textbf{5.50}
& \textbf{14.41} & \textbf{6.43}
& \textbf{9.09} & \textbf{4.98}
& \textbf{11.08} & \textbf{5.02} \\

\bottomrule
\end{tabular}

\label{tab:sota}
\end{table*}

\subsection{State-of-the-art Comparison}

As shown in Tab.~\ref{tab:sota}, ViewSAM trained with weak supervision achieves the best overall performance among zero-shot and weakly supervised approaches, with improvements of at least \textbf{+3.36 in CVR-IDF1 and +3.55 in CVR-MA} across in-domain and cross-domain evaluations, while remaining competitive with fully supervised models. Compared with foundation models (i.e., SAM2~\cite{ravisam} initialized with VLM-generated box prompts, and SAM3~\cite{carion2025sam} using object category as prompts), ViewSAM improves both referring understanding and cross-view association. Moreover, CRTracker trained with the same pseudo labels achieves strong performance, \textbf{validating the effectiveness of our pseudo labels}, yet ViewSAM still surpasses it, highlighting the benefit of view-aware semantics.

Meanwhile, the fully supervised variant of ViewSAM further pushes the performance boundary, achieving improvements of at least \textbf{+2.85 in CVR-IDF1 and +6.03 in CVR-MA} across in-domain and cross-domain evaluations. This also confirms that the proposed view-aware design generalizes well across different supervision settings. The qualitative visualization results can be found in \ref{vis-sota}.

\subsection{Ablation Study and Analysis}

\begin{figure*}[t]
    \centering
    \includegraphics[width=\linewidth]{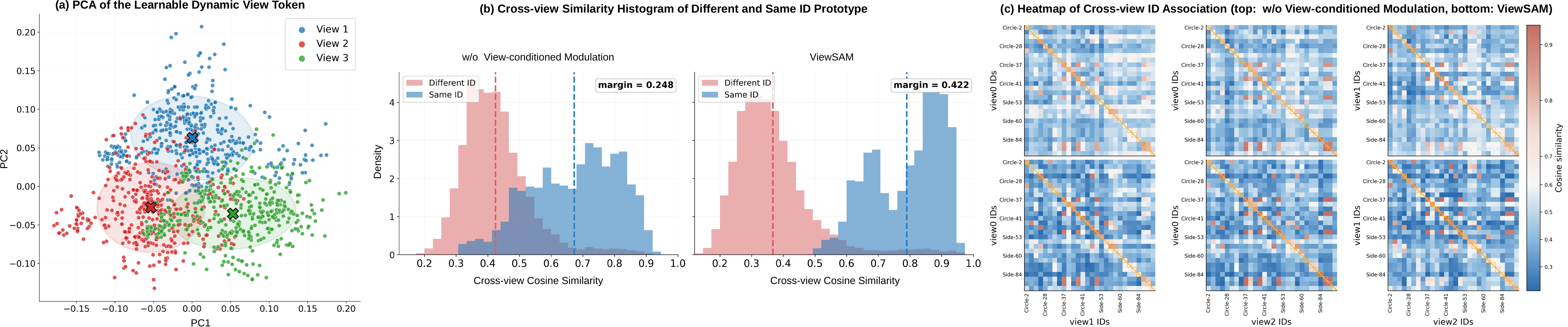}
\caption{Qualitative analysis of view-aware cross-modal semantics. Please zoom in for better visualization. (a) PCA of the view token shows more separable camera-wise clusters. (b) The cross-view similarity distributions exhibit a larger margin between same-ID and different-ID pairs. (c) The association heatmaps demonstrate stronger diagonal dominance and reduced cross-ID interference.}
    \label{vis-view}
\end{figure*}

\begin{figure*}[t]
\centering
\makebox[\linewidth][c]{%
\begin{minipage}[c]{0.5\linewidth}
\centering
\small
\setlength{\tabcolsep}{3pt}
\renewcommand{\arraystretch}{0.9}

\captionof{table}{Contribution of main components.}
\label{tab:ablation_main}
\begin{tabular}{lcc}
\toprule
Method & CVR-IDF1 & CVR-MA \\
\midrule
ViewSAM (full) & \textbf{57.73} & \textbf{43.72} \\
w/o VC-CMA     & 51.43 & 37.54 \\
w/o CG         & 38.19 & 21.30 \\
w/o BAR        & 53.55 & 40.02 \\
w/o CGCT       & 46.50 & 30.49 \\
\bottomrule
\end{tabular}

\captionof{table}{Detailed design ablation results.}
\label{tab:ablation_design}
\begin{tabular}{lcc}
\toprule
Variant & CVR-IDF1 & CVR-MA \\
\midrule
w/o VCCA              & 53.62 & 38.45 \\
\midrule
w/o view modulation   & 52.12 & 35.77 \\
w/o intra-/inter-loss & 50.00 & 33.95 \\
\bottomrule
\end{tabular}

\captionof{table}{Effect of pseudo-label strategies.}
\label{tab:pseudo_quality}
\begin{tabular}{lcc}
\toprule
Pseudo Labels & CVR-IDF1 & CVR-MA \\
\midrule
SAM3 tracklets   & 39.53 & 20.33 \\
+ Assoc.         & 67.87 & 51.61 \\
+ Assoc. + Bi-RP & \textbf{77.96} & \textbf{68.44} \\
\bottomrule
\end{tabular}

\end{minipage}%
\begin{minipage}[c]{0.5\linewidth}
\centering
\includegraphics[width=\linewidth]{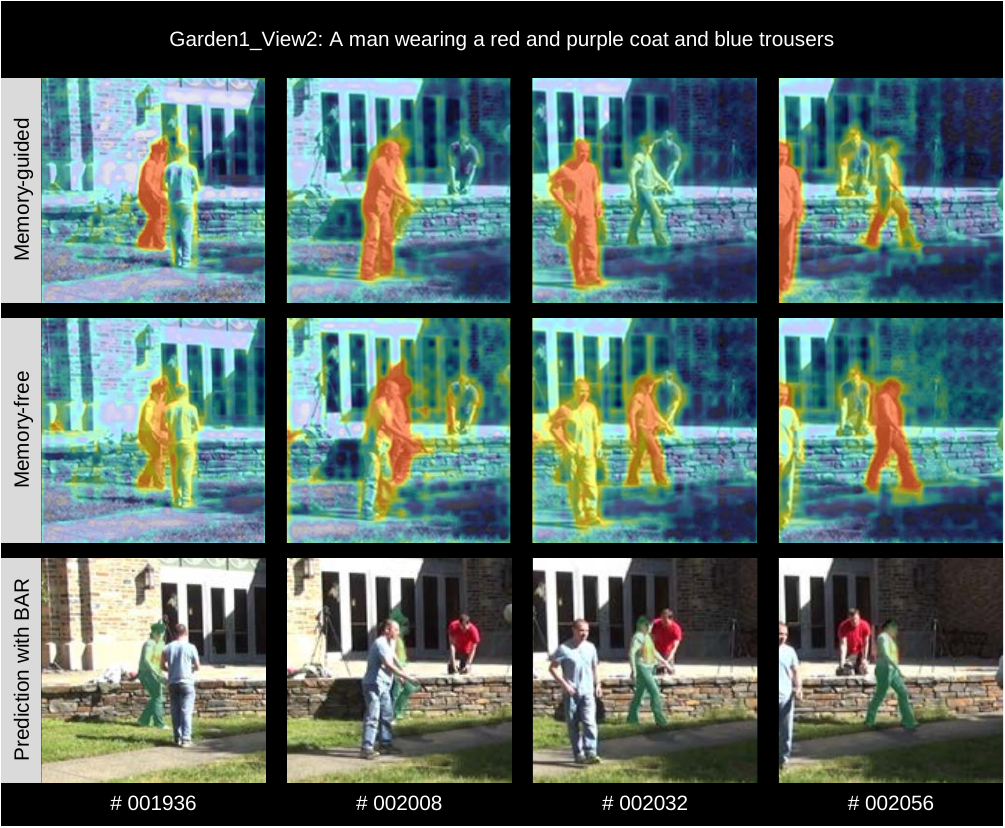}
\captionof{figure}{Visualizations on the effect of Bias-aware Recalibration module, which helps the model to re-focuse on targets that better align with the referring expression by leveraging the memory-free features.}
\label{vis-bar}
\end{minipage}%
}
\end{figure*}

\noindent
\textbf{Impact of view-aware cross-modal semantics.}
We analyze our core contribution from two aspects: (1) the learnable dynamic view token in the VC-CMA and (2) the view-conditioned modulation in the CGCT. Tab.~\ref{tab:ablation_design} shows that removing either design (i.e., w/o VCCA and w/o view modulation) leads to a clear drop, suggesting that the view-aware semantic learning helps reduce the gap between view-variant visual appearances and view-invariant textual semantics for CRMOT. To gain more insights, we present some qualitative results in Fig.~\ref{vis-view}. As shown in Fig.~\ref{vis-view}(a), the dynamic view token yields structured view-wise clusters in the PCA space, indicating that it has learned discriminative view-specific priors as intended. Building upon them, the view-conditioned modulation further aligns instance features across views while preserving identity discriminability. Accordingly, Fig.~\ref{vis-view}(b) shows that the margin between same-/different-ID similarities increases substantially, from 0.248 to 0.422. This is also evident in Fig.~\ref{vis-view}(c), where the association matrix exhibits a clearer diagonal pattern with weaker off-diagonal interference, reflecting more reliable cross-view ID matching.

\noindent
\textbf{Impact of main components and design details.}
As shown in Tab.~\ref{tab:ablation_main}, removing the View-conditioned Cross-Modal Alignment while retaining text prompts (w/o VC-CMA) leads to clear degradation, highlighting the importance of view conditioning for cross-modal alignment. Removing the Candidate Generator (w/o CG) causes a dramatic performance drop, indicating that reliable multi-object proposals are essential for extending SAM2 from single object tracking to multi-object tracking. In contrast, removing the Bias-aware Recalibration (w/o BAR) results in a consistent decline, as also reflected in Fig.~\ref{vis-bar}. Notably, the Consistency-guided Cross-view Tracking head is crucial for cross-view association: removing it (w/o CGCT) significantly degrades performance. As detailed in Tab.~\ref{tab:ablation_design}, discarding consistency objectives also leads to clear degradation.

\noindent
\textbf{Quality of cross-view pseudo labels.}
We select a proxy subset (\textasciitilde 25\% of all samples), where samples are selected based on the similarity of their referring to those in the original annotated training data. Hence, GTs are available for this subset to enable direct evaluation using CRMOT metrics. As shown in Tab.~\ref{tab:pseudo_quality}, affinity-guided association (Assoc.) significantly improves CVR-IDF1 from \textbf{39.53} to \textbf{67.87}, indicating enhanced identity consistency across views. Further introducing bi-directional re-prompting (Bi-RP) boosts both CVR-IDF1 to \textbf{77.96} and CVR-MA to \textbf{68.44}, demonstrating more accurate and reliable cross-view association. More qualitative results are provided in \ref{vis-pseudo}.

\section{Conclusion}
We present the first weakly supervised CRMOT framework using only object category labels. Built on SAM2, ViewSAM leverages view-aware semantics to achieve SOTA performance under weak supervision with \textasciitilde10\% extra parameters and remains competitive with fully supervised methods. Notably, this paradigm naturally supports scalable learning from abundant weakly labeled data.

\bibliographystyle{unsrt}
\bibliography{ref}


\appendix

\section{Qualitative Results for Comparing with SOTAs}\label{vis-sota}

\begin{figure*}[]
    \centering
    \includegraphics[width=\linewidth]{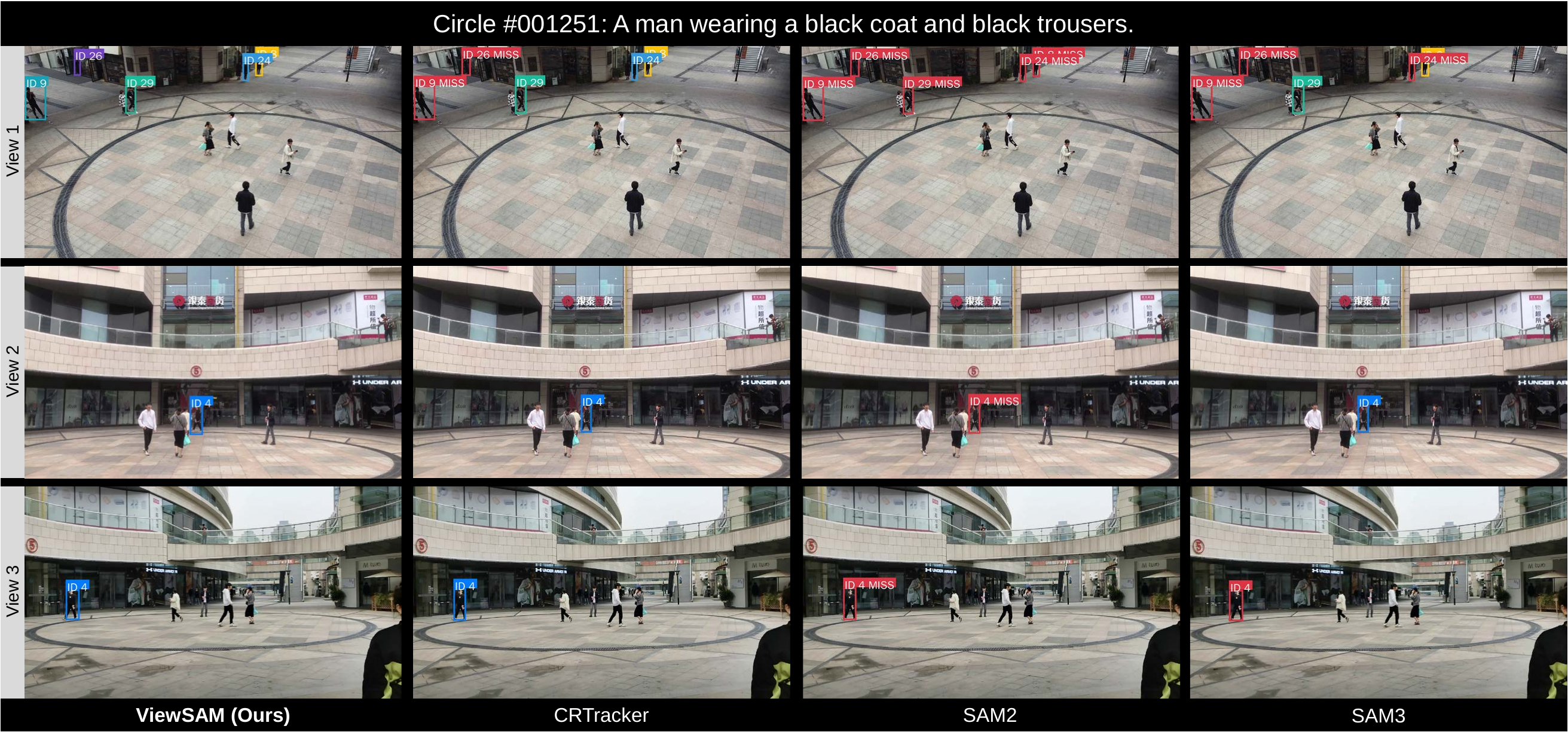}
    \includegraphics[width=\linewidth]{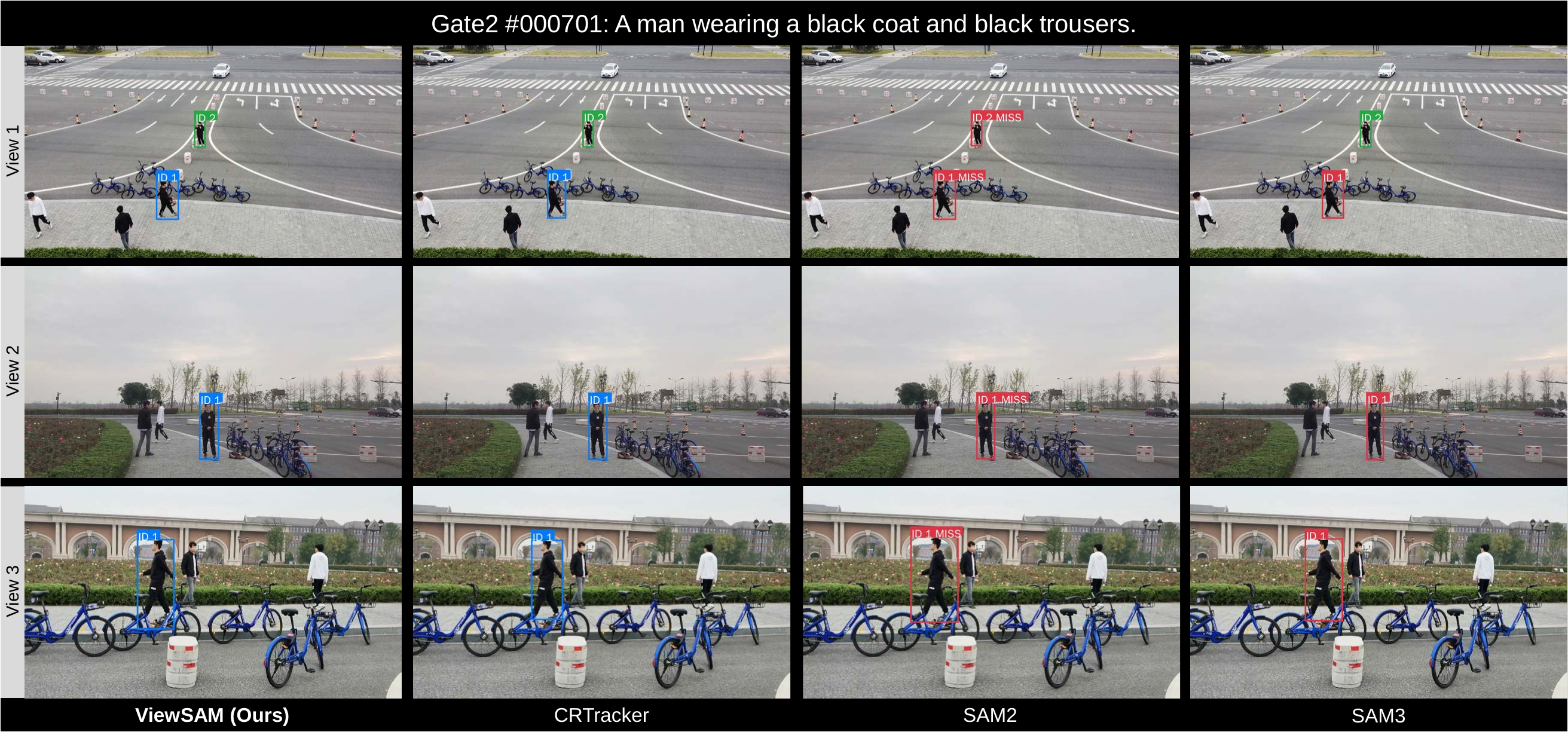}
    \includegraphics[width=\linewidth]{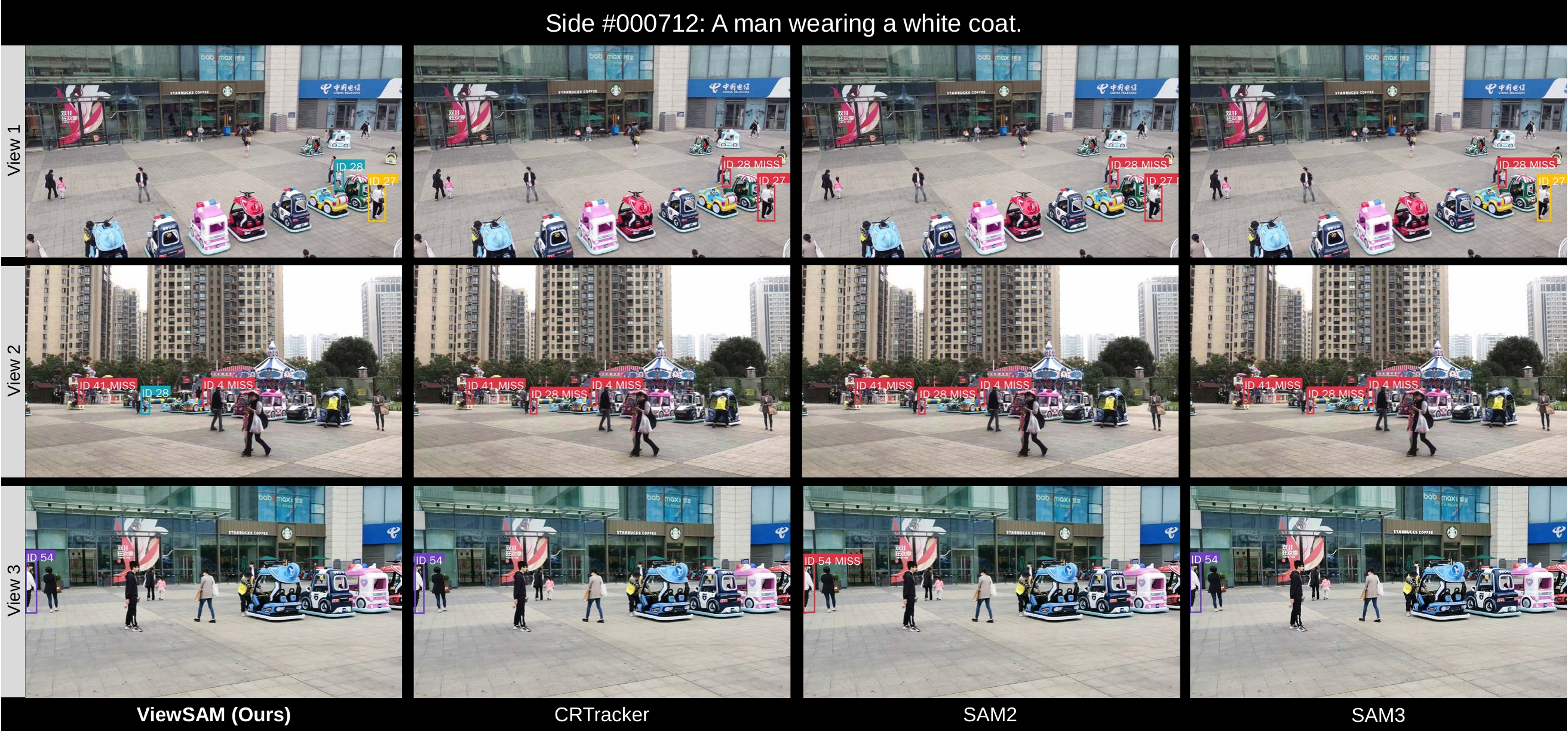}
\caption{Visualization of comparison results on the in-domain scenes.}
    \label{vis-cmp-in}
\end{figure*}

\begin{figure*}[]
    \centering
    \includegraphics[width=0.97\linewidth]{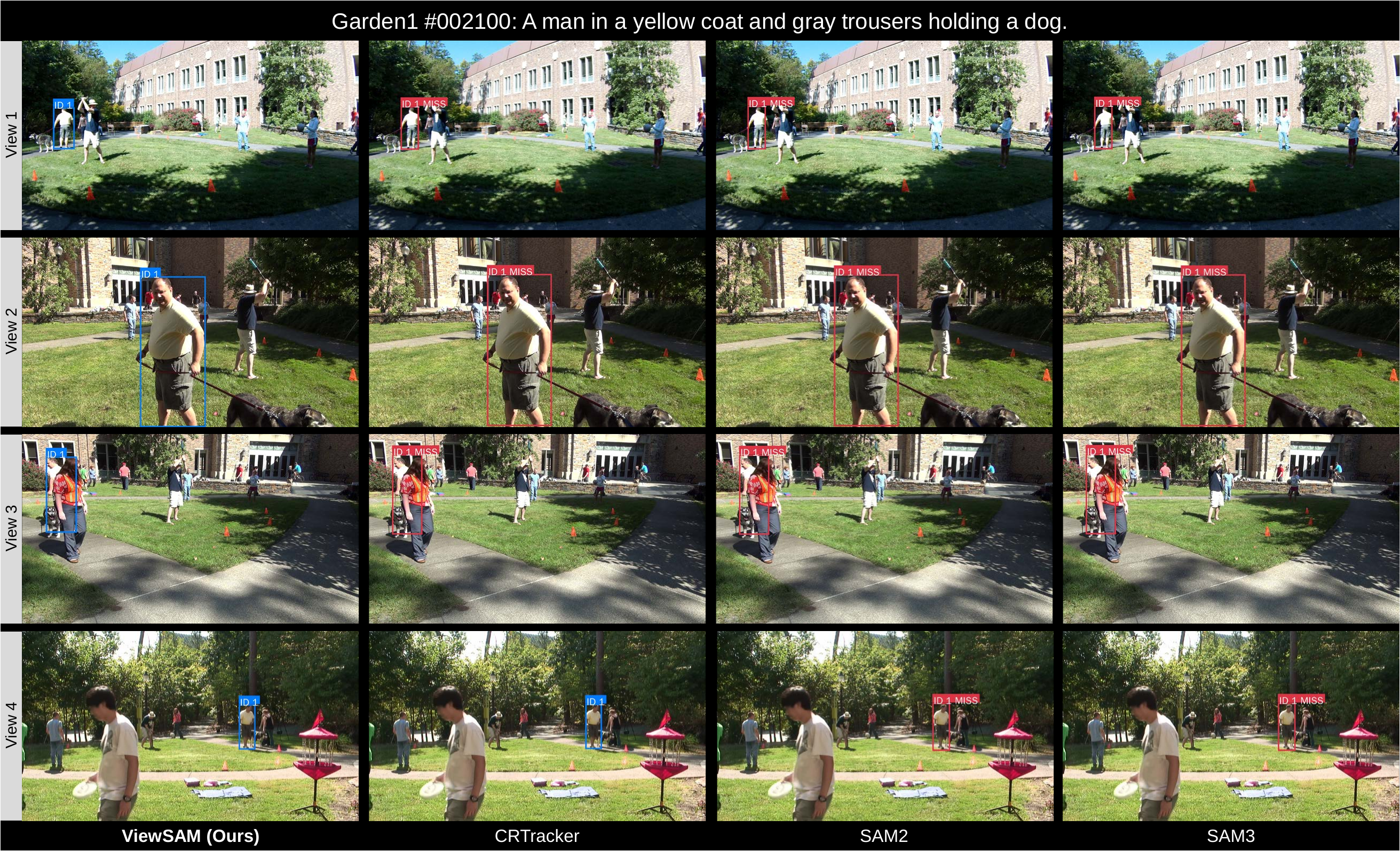}
    \includegraphics[width=0.97\linewidth]{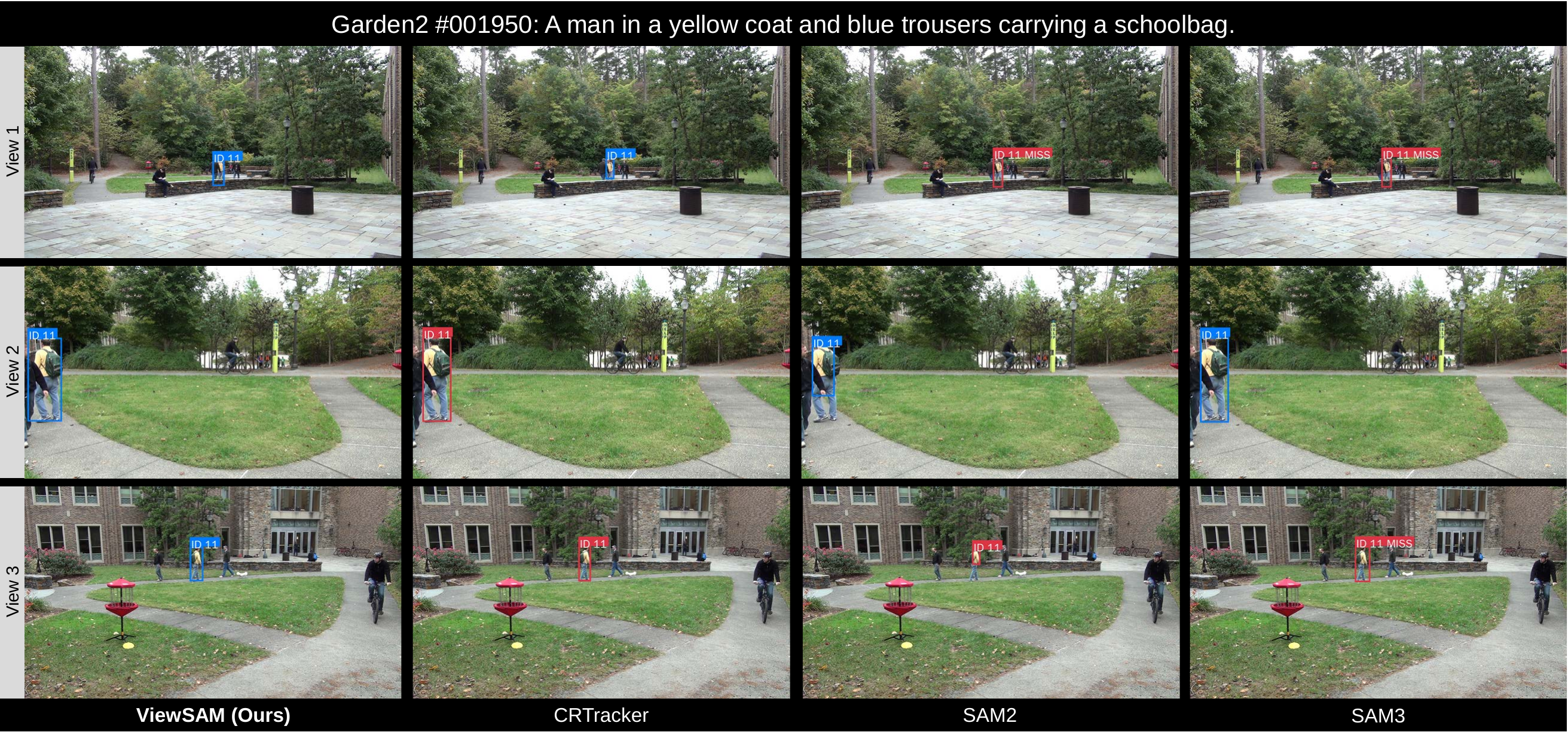}
    \includegraphics[width=0.97\linewidth]{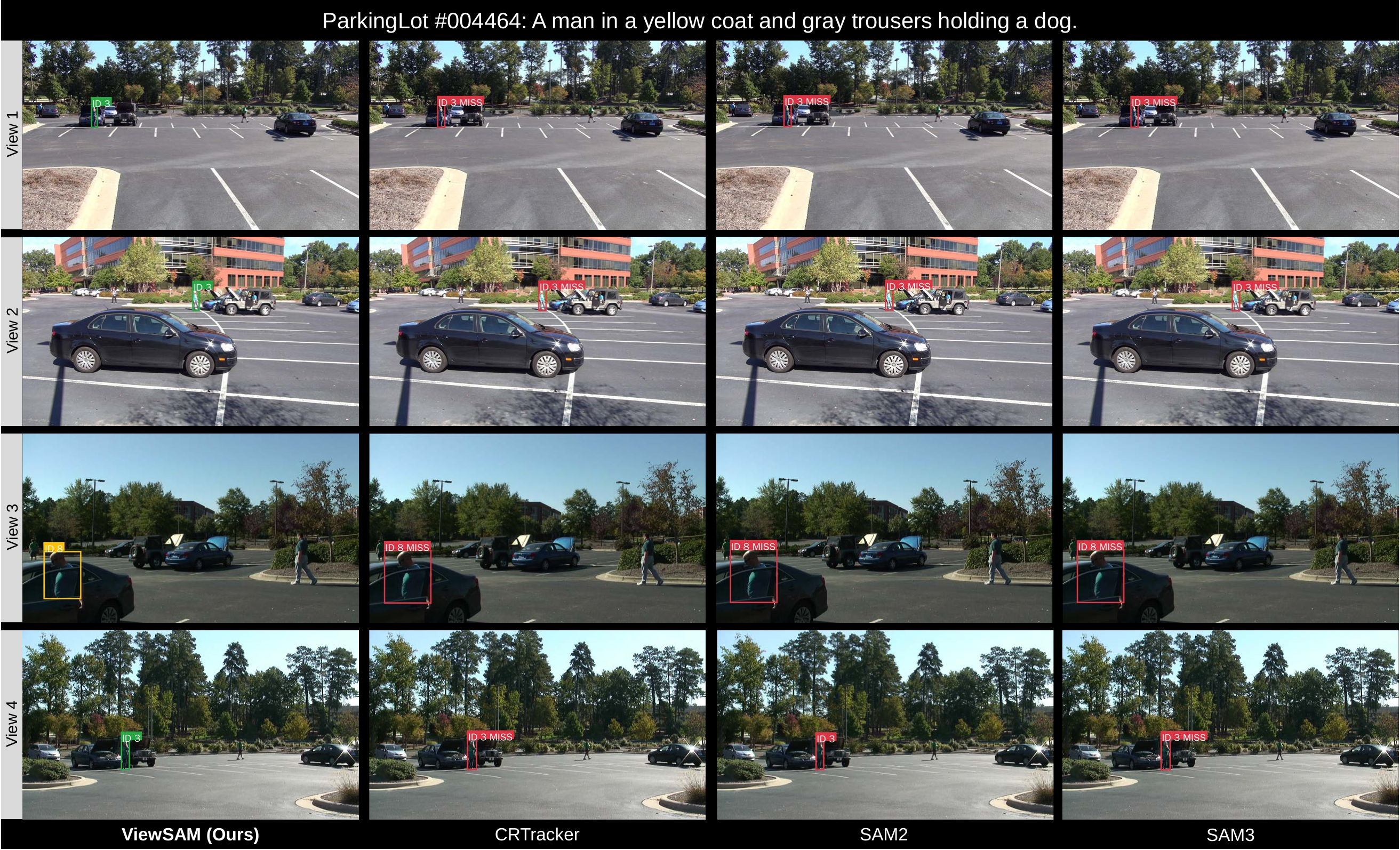}
\caption{Visualization of comparison results on the cross-domain scenes.}
    \label{vis-cmp-cross}
\end{figure*}

To provide intuitive insights beyond quantitative comparisons, we present qualitative results of ViewSAM against representative state-of-the-art methods. As illustrated in Fig.~\ref{vis-cmp-in} and Fig.~\ref{vis-cmp-cross}, we visualize cross-view tracking performance under challenging scenarios, including severe occlusion, appearance ambiguity, and large viewpoint variations. Existing methods often suffer from drift to distractors, inaccurate grounding of referring expressions, or inconsistent identity assignment across views.

In contrast, ViewSAM achieves more precise localization and preserves identity consistency across cameras, even in the presence of heavy occlusions and significant viewpoint changes. This highlights the advantage of incorporating view-aware cross-modal semantics to effectively bridge view-variant visual observations with view-invariant textual semantics, leading to more robust cross-view reasoning and association.

Notably, targets with different IDs are annotated in distinct colors for clarity, while \textbf{red indicates incorrect predictions}. Specifically, \emph{MISS} denotes missed detections, whereas the remaining red cases correspond to identity assignment errors, such as ID switches or incorrect cross-view associations.

\section{Qualitative Results of Cross-view Pseudo Labels}\label{vis-pseudo}

\begin{figure*}[]
    \centering
    \includegraphics[width=\linewidth]{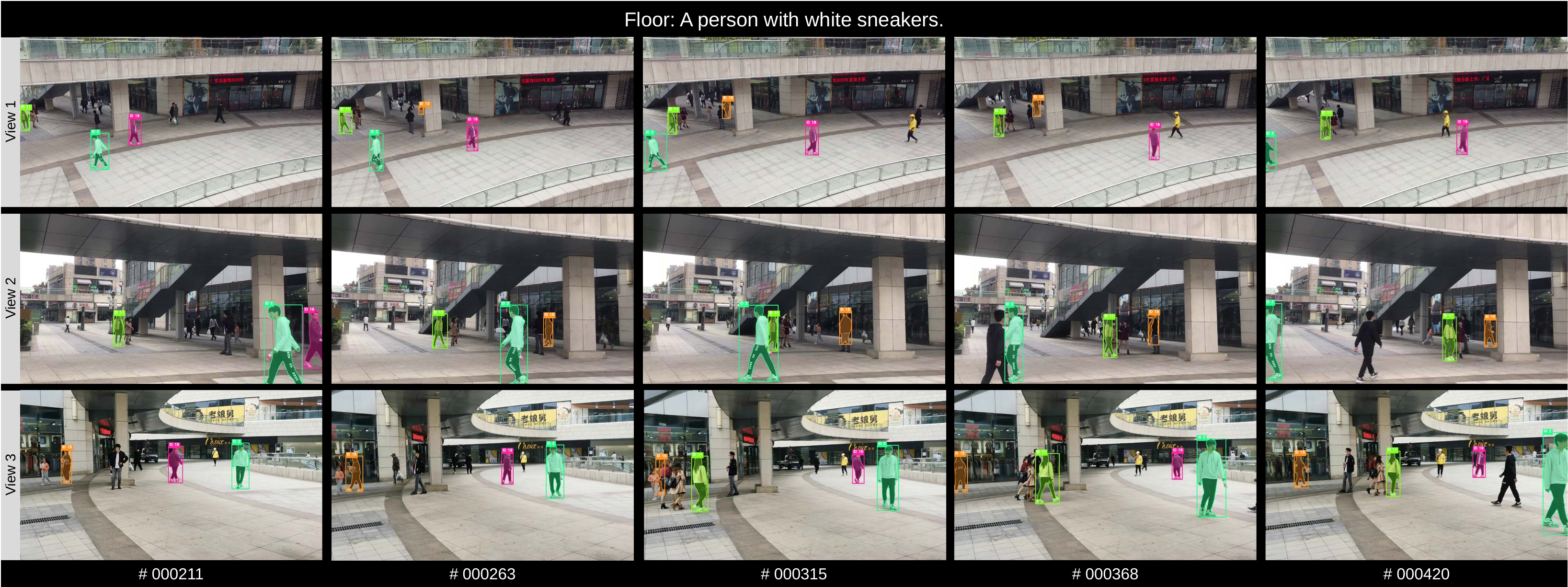}
    \includegraphics[width=\linewidth]{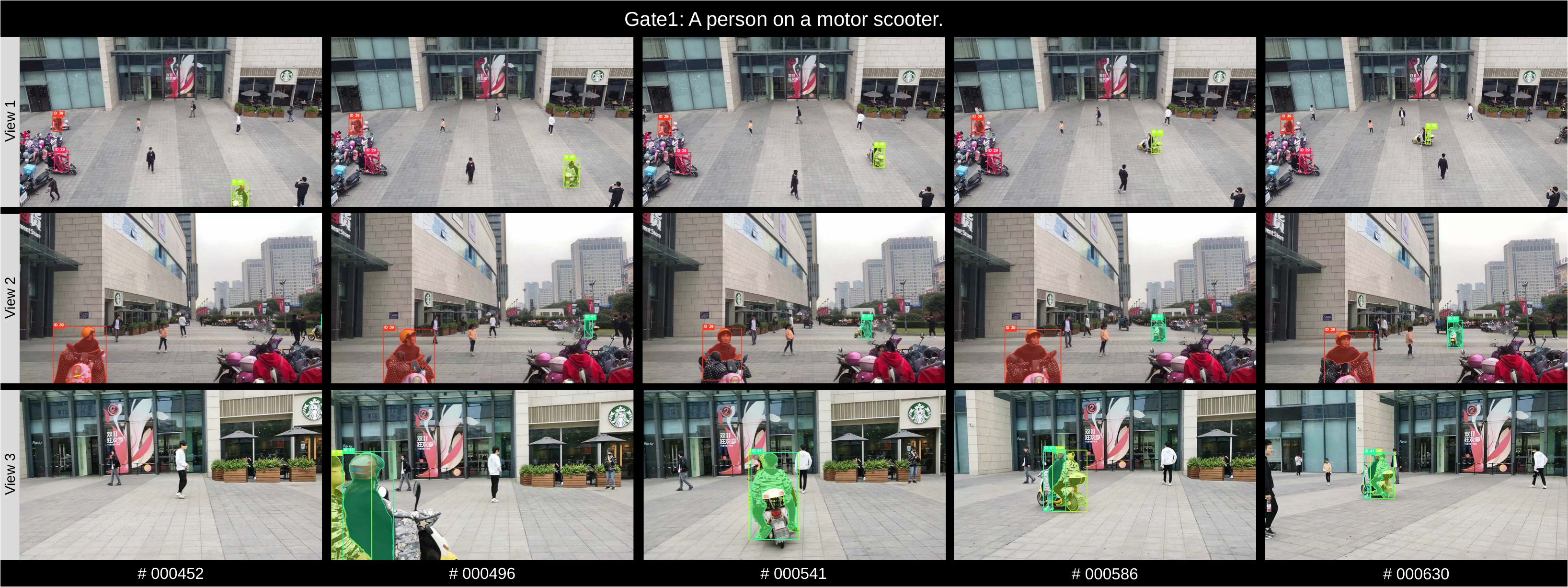}
\caption{Visualization of generated cross-view pseudo labels for CRMOT, illustrating temporal coherence, semantic consistency, and cross-view identity alignment.}
    \label{vis-pseudo-cases}
\end{figure*}

We further visualize the generated cross-view pseudo labels to better understand their role in weakly supervised learning. As shown in Fig.~\ref{vis-pseudo-cases}, the pseudo labels exhibit strong temporal coherence, semantic consistency, and cross-view identity alignment. Specifically, objects maintain stable spatial trajectories across consecutive frames, indicating that the generated tracklets are temporally smooth and robust to motion variations. Moreover, their semantic consistency across views suggests that the pseudo labels capture meaningful object-level cues rather than view-specific artifacts. Importantly, identities are well aligned across different camera views, demonstrating the effectiveness of our cross-view association and re-prompting strategy in mitigating identity ambiguity under weak supervision.

In addition, the pseudo labels remain reliable in challenging scenarios, such as occlusions, scale variations, and viewpoint changes, where naïve applications of foundation models often fail. This highlights the advantage of our affinity-guided cross-view re-prompting mechanism in refining noisy predictions and enforcing structural consistency across views.

Overall, these results demonstrate that our pseudo-label generation pipeline produces high-quality supervision signals that are temporally coherent, semantically consistent, and structurally aligned across views, thereby providing a strong foundation for training the downstream ViewSAM model under the weakly supervised setting.

\section{Details of Main Components}
\label{main-components}

To complement Sec.~\ref{s3-3}, we provide additional details on the main components of ViewSAM. For consistency with the main paper, this section follows the same order as the inference pipeline: \textbf{View-conditioned Cross-modal Alignment (VC-CMA)}, \textbf{Candidate Generator}, \textbf{Bias-aware Recalibration (BAR)}, and \textbf{Consistency-guided Cross-view Tracking Head (CGCT)}. Here, we emphasize only on the architecture and optimization of these modules.

\subsection{View-conditioned Cross-modal Alignment}

As introduced in Sec.~\ref{s3-3}, VC-CMA is designed to inject view-aware cross-modal semantics into the SAM2-based video tracker. It is implemented as a lightweight adaptation module~\cite{houlsby2019parameter} and inserted into selected fusion stages. At each stage, compressed visual tokens interact with textual tokens through view-conditioned cross-attention, where a dynamic View Token provides view-dependent priors for cross-modal alignment.

The dynamic View Token serves as a conditioning signal rather than an independent prediction head. It does not directly produce localization or identity outputs; instead, it modulates visual--text representations to improve their robustness to viewpoint-induced appearance variations. In this way, VC-CMA bridges the gap between view-variant visual observations and the relatively view-invariant semantics of referring expressions.

Since VC-CMA is an adaptation module within the SAM2 mask decoding pipeline, it is trained through the standard mask supervision used by SAM2. Given the predicted mask $\hat M_t$ and the pseudo target mask $M_t$, we optimize VC-CMA using a combination of focal loss and Dice loss:
\begin{equation}
\mathcal{L}_{\mathrm{VC\mbox{-}CMA}}
=
\mathcal{L}_{\mathrm{focal}}(\hat M_t, M_t)
+
\mathcal{L}_{\mathrm{dice}}(\hat M_t, M_t).
\end{equation}

This objective encourages the view-conditioned cross-modal features to improve mask prediction under weak supervision. Through end-to-end optimization, VC-CMA learns view-aware visual--text alignment that is beneficial for robust referring segmentation across different camera views.

\subsection{Candidate Generator}

The Candidate Generator is distilled from APTM~\cite{yang2023towards} and extends SAM2 from single-object tracking to referring multi-object localization. Specifically, we transfer proposal-level localization knowledge by distilling the distribution of proposal quality from APTM, where both proposal confidence and relative ranking are used as supervision signals. This enables the model to inherit strong object localization priors without relying on manual annotations. To further accommodate the weakly supervised CRMOT setting, the model is trained using ID-agnostic single-view pseudo labels generated in Sec.~\ref{s3-3}.

Given the view-aware features produced by VC-CMA, the Candidate Generator predicts a set of candidate boxes, which are subsequently used as box prompts for SAM2 mask decoding. Following proposal-based referring localization paradigms, it consists of two branches: a proposal generation branch for dense object localization, and a proposal--referring matching branch for measuring compatibility with the referring expression.

\noindent
\textbf{Proposal generation branch.}
The proposal branch adopts a dense prediction formulation with an objectness tower and a regression tower, predicting the center heatmap, centerness, box size, and center offset, respectively. Let $\hat H$, $\hat C$, $\hat W$, and $\hat O$ denote the predicted center heatmap, centerness map, box size, and center offset. The objective is defined as
\begin{equation}
\mathcal{L}_{\mathrm{prop}}
=
\lambda_{\mathrm{ctr}}\mathcal{L}_{\mathrm{focal}}(\hat H,H)
+
\lambda_{\mathrm{ctrness}}\mathcal{L}_{\mathrm{focal}}(\hat C,C)
+
\lambda_{\mathrm{wh}}\mathcal{L}_{\mathrm{s-L1}}(\hat W,W)
+
\lambda_{\mathrm{off}}\mathcal{L}_{\mathrm{s-L1}}(\hat O,O),
\end{equation}
where $\mathcal{L}_{\mathrm{focal}}$ is the sigmoid focal loss and $\mathcal{L}_{\mathrm{s-L1}}$ is the Smooth-$L_1$ loss.

\noindent
\textbf{Proposal--referring matching branch.}
To bridge proposal localization and language grounding under weak supervision, each proposal is assigned a matching score with respect to the referring expression. Specifically, proposal features are extracted from multi-scale visual representations, fused with the referring embedding and box geometry, and mapped to proposal-level matching logits. Let $s_i$ denote the matching logit of proposal $i$. 

We construct soft supervision targets from pseudo labels:
\begin{equation}
y_i
=
\mathrm{clip}
\left(
\frac{\mathrm{IoU}_i-\tau_{\mathrm{neg}}}{\tau_{\mathrm{pos}}-\tau_{\mathrm{neg}}},
0,1
\right),
\end{equation}
where $\mathrm{IoU}_i$ denotes the overlap between proposal $i$ and the pseudo target, and $\tau_{\mathrm{pos}}$ and $\tau_{\mathrm{neg}}$ are positive and negative thresholds. The matching loss is
\begin{equation}
\mathcal{L}_{\mathrm{match}}
=
\lambda_{\mathrm{match}}\mathcal{L}_{\mathrm{gbce}}(\{s_i\},\{y_i\}),
\end{equation}
where $\mathcal{L}_{\mathrm{gbce}}$ denotes a group-balanced binary cross-entropy loss that normalizes positive and negative proposal groups separately.

To further enlarge the margin between valid and invalid proposals, we introduce an auxiliary ranking objective that also facilitates distribution distillation from APTM:
\begin{equation}
\mathcal{L}_{\mathrm{rank}}
=
\frac{1}{|\mathcal{P}||\mathcal{N}|}
\sum_{i\in\mathcal{P}}
\sum_{j\in\mathcal{N}}
\max(0,m-(s_i-s_j)),
\end{equation}
where $\mathcal{P}$ and $\mathcal{N}$ denote the positive and negative proposal sets derived from pseudo labels. The final objective is
\begin{equation}
\mathcal{L}_{\mathrm{cand}}
=
\mathcal{L}_{\mathrm{prop}}
+
\mathcal{L}_{\mathrm{match}}
+
\lambda_{\mathrm{rank}}\mathcal{L}_{\mathrm{rank}}.
\end{equation}

In practice, the Candidate Generator serves as the bridge between view-aware representations and SAM2 mask decoding, transferring proposal distribution knowledge distilled from APTM into reliable box prompts, and enabling the transition from single-object tracking to referring multi-object localization under weak supervision.

\subsection{Bias-aware Recalibration}

BAR is introduced to mitigate tracking bias caused by error accumulation in memory-guided tracking in SAM2. Memory-guided decoding may amplify early mistakes and gradually drift to distractors, especially under occlusion or severe appearance ambiguity. To address this issue, BAR learns to detect when the memory-guided prediction becomes unreliable by comparing it with a memory-free prediction that does not use the tracking history stored in the Memory Bank.

Specifically, for each target candidate, we compute two masks. The first mask is predicted from the memory-guided branch:
\begin{equation}
\hat M^{\mathrm{b}}_t =
\mathcal{D}_{\mathrm{dec}}(\mathcal{F}_{\mathrm{mem}}, \rho) > 0,
\end{equation}
where $\mathcal{F}_{\mathrm{mem}}$ denotes the memory-enhanced features and $\rho$ is the prompt. This prediction corresponds to the standard SAM2 decoding process conditioned on the Memory Bank.

To obtain an unbiased reference, we further compute a memory-free mask:
\begin{equation}
\hat M^{\mathrm{u}}_t =
\mathcal{D}_{\mathrm{dec}}(\mathcal{F}, \rho) > 0,
\end{equation}
where $\mathcal{F}$ denotes the features without memory guidance. Since this branch does not access previous tracking context, its prediction is determined mainly by the current-frame visual content and the referring prompt.

Given the two binary masks at frame $t$, we define the bias supervision label as:
\begin{equation}
y_t =
\begin{cases}
1, & \text{if } \hat M^{\mathrm{u}}_t \cap \hat M^{\mathrm{b}}_t = \emptyset,\\
0, & \text{otherwise}.
\end{cases}
\end{equation}
Here, $y_t=1$ indicates that the memory-guided and memory-free branches segment different objects, suggesting that the memory-guided prediction may have drifted to a distractor. We supervise the predicted bias probability $\hat b_t$ using a standard cross-entropy loss:
\begin{equation}
\mathcal{L}_{\mathrm{bar}}
=
-\frac{1}{T}
\sum_{t=1}^{T}
\left[
y_t \log(\hat b_t)
+
(1-y_t)\log(1-\hat b_t)
\right].
\end{equation}

Through this supervision, BAR learns to identify frames where memory guidance becomes unreliable and adaptively recalibrates the final mask prediction toward the memory-free branch when necessary.

\subsection{Consistency-guided Cross-view Tracking Head}

The CGCT head is responsible for cross-view identity association. It is built upon an OSNet \cite{zhou2019omni} to capture appearance features for robust representation learning. As described in Sec.~\ref{s3-3}, it first extracts object-level representations via masked pooling over multi-scale features and projects them into a shared embedding space for identity matching. To alleviate appearance discrepancies across cameras, the dynamic View Token is injected through view-conditioned FiLM modulation, producing embeddings that are view-aware during feature adaptation while encouraging view-invariant representations for identity association.

The core supervision of CGCT follows a consistency-guided formulation that explicitly models both intra-view and inter-view identity structures. Let $z_{t,i}^{k}$ denote the embedding of instance $k$ at frame $t$ and view $i$, $\bar z_g^i$ the prototype of identity $g$ in view $i$, and $\bar z_g$ the global prototype aggregated across views. The training objective is
\begin{equation}
\small
\begin{aligned}
\mathcal{L}_{\mathrm{cgct}}
=
\sum_g
\Bigg[
\sum_i \sum_{(t,k)\in\mathcal{T}_g^i}
\left\|
z_{t,i}^{k} - \bar z_g^i
\right\|_2^2
+
\lambda
\sum_i
\left\|
\bar z_g^i - \bar z_g
\right\|_2^2
\Bigg],
\end{aligned}
\end{equation}
where the first term enforces intra-view consistency by compacting embeddings belonging to the same trajectory within each camera view, while the second term enforces inter-view consistency by aligning view-specific prototypes toward a shared global identity prototype.

In contrast to conventional association heads that rely solely on appearance similarity, CGCT explicitly models a hierarchical consistency structure, capturing temporal coherence within each view and identity agreement across views. This formulation is particularly suited to the weakly supervised setting, where explicit cross-view identity annotations are unavailable. By leveraging pseudo supervision and consistency constraints, CGCT learns to suppress view-induced discrepancies and establish robust identity associations across cameras.

\section{Limitations}\label{limi}
Despite its effectiveness, our framework has several limitations. First, it relies on pseudo labels generated in Stage 1, whose quality is bounded by the performance of SAM3 and cross-view association; errors may propagate to downstream training, especially in crowded scenes or under severe occlusion. Moreover, the weak supervision setting still assumes synchronized multi-view videos and coarse-grained category labels, which may limit applicability in more unconstrained scenarios. Meanwhile, the pipeline depends on multiple external components (e.g., SAM3, ReID models, and MLLMs), increasing system complexity and computational cost. Finally, while ViewSAM models view-aware cross-modal semantics, it remains challenged by ambiguous referring expressions and complex multi-object interactions, leaving room for improving language grounding and temporal reasoning.

\section{Impact}\label{impact}
This work contributes to reducing the reliance on dense annotations for cross-view referring multi-object tracking, which may facilitate scalable research and applications in multi-camera perception systems such as intelligent surveillance and embodied AI. However, the ability to track objects across views based on language descriptions also raises potential privacy and misuse concerns, particularly in surveillance scenarios. Moreover, biases inherited from foundation models or pseudo-label generation may affect fairness and reliability. We emphasize that this work is intended for research purposes, and any real-world deployment should comply with ethical standards, privacy regulations, and appropriate human oversight.



\end{document}